\documentclass[runningheads,a4paper]{llncs}

\usepackage{amssymb}
\usepackage{graphicx}
\usepackage{epstopdf}
\usepackage{multicol}
\usepackage{url}
\urldef{\mailsa}\path|{jaga.math23@gmail.com, anantarj11593@gmail.com, animesh2007.1813@rediffmail.com, dr.samarjitkar@gmail.com}|
\urldef{\mailsb}\path||
\urldef{\mailsc}\path||    
\newcommand{\keywords}[1]{\par\addvspace\baselineskip
\noindent\keywordname\enspace\ignorespaces#1}

\begin{document}

\mainmatter  

\title{An extended  Multi Attributive Border Approximation area Comparison using interval type-2 trapezoidal fuzzy numbers}

\titlerunning{Extended  MABAC using interval type-2 trapezoidal fuzzy numbers}

%
%
\author{Jagannath Roy%
\and Ananta Ranjan \and Animesh Debnath \and Samarjit Kar\thanks{Corresponding author}\\}
\authorrunning{J. Roy, A.Ranjan, A. Debnath and S Kar}

\institute{Department of Mathematics, National Institute of Technology,\\ Durgapur-713209, India\\
\mailsa\\
\mailsb\\
\mailsc\\}

%
%

\toctitle{Extended  MABAC using interval type-2  trapezoidal fuzzy numbers}
\tocauthor{J. Roy, A.Ranjan and A. Debnath}
\maketitle

\begin{abstract}
In this paper, we attempt to extend Multi Attributive Border Approximation area Comparison (MABAC) approach for multi-attribute decision making (MADM) problems  based on type-2 fuzzy sets (IT2FSs). As a special case of IT2FSs interval type-2 trapezoidal fuzzy numbers (IT2TrFNs) are adopted here to deal with uncertainties present in many practical evaluation and selection problems. A systematic description of MABAC based on IT2TrFNs is presented in the current study. The validity and feasibility of the proposed method are illustrated by a practical example of selecting the most suitable candidate for a software company which is heading to hire a system analysis engineer based on few attributes. Finally, a comparison with two other existing MADM methods is described.
\end{abstract}

\section{Introduction}

Multi-Attributive Border Approximation area Comparison (MABAC) developed by Pamu{\v{c}}ar and {\'C}irovi{\'c} \cite{pamucar15} is one of the recent approaches to multi-attribute decision making (MCDM/MADM) problems. MABAC method has attracted researchers due to its inherent characteristics: simple computation and stability in solution. The basic principle used in MABAC is that it divides the performances of each criteria/attribute function into Upper Approximation Area (UAA) containing ideal alternatives and Lower Approximation Area (LAA) containing anti-ideal alternatives. In other words this method provides a direct relextion of
the relative strength and weakness of an alternative over others' according to each criteria. In classical MABAC \cite{pamucar15} method, the performance rating and criteria weights were represented by crisp/deterministic numerical values. But human judgment, preference values and criteria weights are often imprecise and ambiguous in nature and cannot be represented by crisp numbers in real-life problems. In response to the uncertainty inherent from decision makers' subjective judgment, Zadeh \cite{zadeh65}, first,  introduced fuzzy set theory (FST). Further extensions of FST done by many authors. Atanassov \cite{atan86} defined intuitionistic fuzzy set(IFS) as a generalization of fuzzy sets to formulate the non-determinacy which occurs due to hesitation of decision makers in the system. Further generalization of IFS as interval-valued intuitionistic fuzzy set (IVIFS) by Atanassov and Gargov \cite{atan89} where membership and non-membership are introduced to deal with impreciseness in decision making. Xue \cite{xu07} used IVIFS in MCDM problems after developing some aggregation operators for IVIFS. In this regard, few extensions of MABAC method could be found in the literature. Recently, Peng and Yang \cite{peng16} developed Pythagorean Fuzzy Choquet Integral Based MABAC Method for Multiple Attribute Group Decision Making. Xue et al. \cite{xue16} defined interval-valued intuitionistic fuzzy MABAC approach and used for material selection with incomplete weight information. Bo{\v{z}}ani{\'c} et al. \cite{bozani16} used the fuzzy AHP-MABAC hybrid model in ranking potential locations for preparing laying-up positions. Roy et al. \cite{roy16} extended MABAC model using rough approximation and rough numbers and developed a hybrid MADM method for assessing medical tourism sites in India.\\

Recent researchers are applying Fuzzy Type-2 sets having fuzzy-based membership function, dealing with benefit and risk under uncertain environment in a better way. Type-2 fuzzy sets (T2FS) \cite{mendel02} are extension of ordinary type-1 fuzzy sets where traditional Type-1 fuzzy sets fails to deal with uncertainties, due to crisp nature of their membership function. Lee and Chen \cite{lee08} present a new method for handling fuzzy multiple criteria hierarchical group decision-making problems based on arithmetic operations and fuzzy preference relations of interval type-2 fuzzy sets. Chen and Lee \cite{chen10a} also presented an interval type-2 TOPSIS method to handle fuzzy MCDM problem. Chen and Lee \cite{chen10b} suggested a method to handle MCDM problems based on ranking values and arithmetic operation of interval type-2 fuzzy set. Wang et al. \cite{wang12} proposed a method for MAGDM based on interval type-2 fuzzy environment. Chen et al. \cite{chen12} presented a new method to handle fuzzy MAGDM problems based on ranking interval type-2 fuzzy sets. Celik et al. \cite{celik13} proposed an integrated novel type-2 fuzzy MCDM method to improve customer satisfaction in public transportation in Istanbul. Chen and Wang \cite{chen13a} applied interval type-2 fuzzy set in fuzzy decision making. A linear assignment method for multiple-criteria decision analysis with interval type-2 fuzzy sets is discussed in Chen \cite{chen13b}.  Kar and Chatterjee \cite{kar15} applied  ranking based interval type-2 fuzzy set exploring the risk factors and ranking supplier companies and presented an empirical study. Despite various methods developed based on type-2 fuzzy sets, these methods have their own shortcomings-large number of computations, i.e., computation complexity. But MABAC has simple computation steps and provides stable solutions. With these considerations, the paper extends the MABAC method for IT2TrFNs for making pragmatic and reliable decisions in evaluation and selection of the suitable alternative/candidate for a job.

The remainder of this paper is structured as follows. In Section 2, we first introduce some preliminaries on type-2 fuzzy sets and  interval type-2 trapezoidal fuzzy numbers (IT2TrFNs). In Section 3, we extend MABAC method based on IT2TrFNs. Section 4 presents an illustrative example for validation of our proposed model. A comparison of the proposed MADM method with other two methods is presented in section 5. Finally, we conclude the work in Section 6.

\section{Preliminaries}

In this section, we briefly review some definitions of type-2 fuzzy sets and interval type-2 fuzzy sets.

\begin{definition}\emph{\cite{mendel02} 
A type-2 fuzzy set (T2FS) $\tilde{\tilde{A}}$ in the universe of discourse $X$ can be represented by a type-2 membership function $\mu_{A}$ shown as follows:
\begin{equation}
\tilde{\tilde{A}}=\lbrace((x,u), \mu_{A}(x,u))|\forall x\in X ,\forall u\in J_{x}\subseteq [0,1], 0\leq \mu_{A}(x,u) \leq 1\rbrace
\end{equation} 
where $J_{x}$ denotes an interval in $[0, 1]$. Moreover, the type-2 fuzzy set $\tilde{\tilde{A}}$ also can be represented as follows:
\begin{equation}
\tilde{\tilde{A}}=\int_{x\in X}\int_{u\in J_{x}} \mu_{A}(x, u)/(x, u)= \int_{x\in X}\left(\int_{u\in J_{x}} \mu_{A}(x, u)/u\right)/x  
\end{equation} 
where $J_{X}\subseteq [0,1]$ is the primary membership at $x$, and $\int_{u\in J_{x}} \mu_{A}(x, u)/u$ indicates the second membership at $x$. For discrete space, the symbol $\int$ is replaced by $\sum$}.
\end{definition}
\begin{definition}\emph{\cite{lee08}
Let $\tilde{\tilde{A}}$ be a type-2 fuzzy set in the universe of discourse $X$ represented by a type-2 membership function $\mu_{A}(x, u)$. If all $\mu_{A}(x, u)=1$, then $\tilde{\tilde{A}}$ is called an interval type-2 fuzzy set (IT2FS). An IT2FS can be regarded as a special case of the T2FS, which is defined as follows:}
\end{definition}
\begin{equation}
\tilde{\tilde{A}}=\int_{x\in X}\int_{u\in J_{x}} 1/(x, u)= \int_{x\in X}\left(\int_{u\in J_{x}} 1/u\right)/x  
\end{equation} 
It is obvious that the IT2FSs $\tilde{\tilde{A}}$ defined in $X$ is fully determined by the primary membership which is called the footprint of uncertainty (FOU), and can be expressed as follows:
\begin{equation}
FOU\left(\tilde{\tilde{A}}\right)=\bigcup_{x\in X} J_{x}=\bigcup_{x\in X}\lbrace (x, u)| u\ J_{x}\subseteq [0, 1]\rbrace 
\end{equation} 
Because the operations on IT2FS are quite complex, IT2FS is usually considered by studying some simplified versions. Here, we follow the results of Chen \cite{chen13b}, who adopted  interval type-2 trapezoidal fuzzy numbers (IT2TrFNs) for solving MADM problems.

\begin{definition}\emph{\cite{chen13b} 
Let $A^U$ and $A^L$ be two  interval type-2 trapezoidal fuzzy numbers (IT2TrFNs), where the height of a generalized fuzzy number is between zero to one. Let $h_{A}^U$ and $h_{A}^L$ be the heights of $A^U$ and $A^L$, respectively. A TrIT2FS $\tilde{\tilde{A}}$ in the universe of discourse $X$ is defined in the form:}
\begin{equation}
\tilde{\tilde{A}}=[A^U, A^L]=[(a_{1}^U, a_{2}^U, a_{3}^U, a_{4}^U;h_{A}^U), (a_{1}^L, a_{2}^L, a_{3}^L, a_{4}^L;h_{A}^L)] 
\end{equation} 
\end{definition}
where $a_{1}^U, a_{2}^U, a_{3}^U, a_{4}^U;h_{A}^U, a_{1}^L, a_{2}^L, a_{3}^L, a_{4}^L$ and $h_{A}^L$ are all real numbers and which satisfy the inequalities $a_{1}^U\leq a_{2}^U\leq a_{3}^U\leq a_{4}^U$; $a_{1}^L\leq a_{2}^L\leq a_{3}^L\leq a_{4}^L$ and $0\leq h_{A}^L \leq h_{A}^U$. The upper membership function (UMF) $h_{A}^U$ and lower membership function (LMF) $h_{A}^L$ are defined in the following way (see Fig. 1):
\begin{equation}
A^U (x)=\left\{
   \begin{array}{ll}
   \frac{(x-a_{1}^U)h_{A}^U}{a_{2}^U-a_{1}^U}, & \hbox{if $a_{1}^U\leq x\leq a_{2}^U$} \\
   h_{A}^U, & \hbox{if $a_{2}^U\leq x\leq a_{3}^U$} \\
   \frac{(a_{4}^U-x)h_{A}^U}{a_{4}^U-a_{3}^U}, & \hbox{if $a_{3}^U\leq x\leq a_{4}^U$} \\
   0, & \hbox{otherwise}
  \end{array}
  \right.
\end{equation} 
and
\begin{equation}
A^L (x)=\left\{
   \begin{array}{ll}
   \frac{(x-a_{1}^L)h_{A}^L}{a_{2}^L-a_{1}^L}, & \hbox{if $a_{1}^L\leq x\leq a_{2}^L$} \\
   h_{A}^L, & \hbox{if $a_{2}^L\leq x\leq a_{3}^L$} \\
   \frac{(a_{4}^L-x)h_{A}^L}{a_{4}^L-a_{3}^L}, & \hbox{if $a_{3}^L\leq x\leq a_{4}^L$} \\
   0, & \hbox{otherwise}
  \end{array}
  \right.
\end{equation} 

\subsection{Geometric Bonferroni Mean operator for IT2TrFNs}
\begin{definition}\emph{\cite{gong15} 
Let $\tilde{\tilde{A}}_{i}=(A_{i}^U, A_{i}^L)= [(a_{i1}^U, a_{i2}^U, a_{i3}^U, a_{i4}^U; h_{A_{i}}^U), (a_{i1}^L, a_{i2}^L, a_{i3}^L, a_{i4}^L; h_{A_{i}}^L)],$ $(i=1, 2,\cdots, n)$ be a collection of IT2TrFNs and $r, s\geq 0$, then, we call: 
\begin{equation}
TIT2FGBM^{r, s}\left(\tilde{\tilde{A}}_{1}, \tilde{\tilde{A}}_{2}, \cdots, \tilde{\tilde{A}}_{n}\right)=\frac{1}{p+q}\left(\otimes_{i, j=1}^{n}(r\tilde{\tilde{A}}_{i} \oplus s\tilde{\tilde{A}}_{j})\right)^{1/n(n-1)}
\end{equation} 
a interval type-2 trapezoidal Fuzzy Geometric Bonferroni Mean (TIT2FGBM) operator.}
\end{definition}
\begin{theorem}\emph{\cite{gong15} 
Let $\tilde{\tilde{A}}_{i}=(A_{i}^U, A_{i}^L)= [(a_{i1}^U, a_{i2}^U, a_{i3}^U, a_{i4}^U; h_{A_{i}}^U), (a_{i1}^L, a_{i2}^L, a_{i3}^L, a_{i4}^L; h_{A_{i}}^L)]$, $(i=1, 2,\cdots, n)$ be a collection of IT2TrFNs and $r, s\geq 0$, then, the aggregated result by Eq. (14) is also a IT2TrFN, and}
\begin{equation}
TIT2FGBM^{r, s}\left(\tilde{\tilde{A}}_{1}, \tilde{\tilde{A}}_{2}, \cdots, \tilde{\tilde{A}}_{n}\right)=\tilde{\tilde{A}}=(A^U, A^L)
\end{equation}
where
\begin{equation}
\begin{array}{ll}
A^U=\Big(\prod\limits_{i, j=1, i\neq j}^n\left(ra_{i1}^U+sa_{j1}^U\right)^{1/n(n-1)}, \prod\limits_{i, j=1, i\neq j}^n\left(ra_{i2}^U+sa_{j2}^U\right)^{1/n(n-1)},\\ \prod\limits_{i, j=1, i\neq j}^n\left(ra_{i3}^U+sa_{j3}^U\right)^{1/n(n-1)}, \prod\limits_{i, j=1, i\neq j}^n\left(ra_{i4}^U+sa_{j4}^U\right)^{1/n(n-1)}; h_{A}^U=\min\limits_{1\leq i\leq n}(h_{A_{i}}^U) \Big)
\end{array}
\end{equation}
and
\begin{equation}
\begin{array}{ll}
A^L=\Big(\prod\limits_{i, j=1, i\neq j}^n\left(ra_{i1}^L+sa_{j1}^L\right)^{1/n(n-1)}, \prod\limits_{i, j=1, i\neq j}^n\left(ra_{i2}^L+sa_{j2}^L\right)^{1/n(n-1)},\\ \prod\limits_{i, j=1, i\neq j}^n\left(ra_{i3}^L+sa_{j3}^L\right)^{1/n(n-1)}, \prod\limits_{i, j=1, i\neq j}^n\left(ra_{i4}^L+sa_{j4}^L\right)^{1/n(n-1)}; h_{A}^L=\min\limits_{1\leq i\leq n}(h_{A_{i}}^L) \Big)
\end{array}
\end{equation}
\end{theorem}

\subsection{The ranking-based distance function of IT2TrFNs}
\begin{theorem}\emph{\cite{qin15}  
Let $\tilde{\tilde{A}}=[(a_{1}^U, a_{2}^U, a_{3}^U, a_{4}^U;h_{A}^U), (a_{1}^L, a_{2}^L, a_{3}^L, a_{4}^L;h_{A}^L)]$ be a TrIT2FS defined in the universe of discourse $X$. The rank-based distance function between $\tilde{\tilde{A}}$ and $\tilde{\tilde{1}}$ are defined as follows:}
\begin{equation}
\begin{array}{ll}
R_{d}(\tilde{\tilde{A}}, \tilde{\tilde{1}})=1-a_{4}^L-\lambda(a_{1}^L-a_{1}^U+a_{4}^U-a_{4}^L)-\frac{1}{2 h_{A}^U h_{A}^L}[h_{A}^U(\lambda(a_{2}^L-a_{1}^L-a_{2}^U+a_{1}^L)\\-(a_{4}^L-a_{3}^L-a_{2}^L+a_{1}^L)-h_{A}^L(a_{4}^U-a_{3}^U-a_{4}^L+a_{3}^L)]
\end{array}
\end{equation} 
\end{theorem}
\begin{definition}\emph{\cite{qin15} 
Let $\tilde{\tilde{A}}$ and $\tilde{\tilde{B}}$ be two IT2TrFNs. Then the distance between $\tilde{\tilde{A}}$ and $\tilde{\tilde{B}}$ is defined as:}
\begin{equation}
d\left(\tilde{\tilde{A}}, \tilde{\tilde{B}}\right)=\left|R_{d}\left(\tilde{\tilde{A}}, \tilde{\tilde{1}}\right) - R_{d}\left(\tilde{\tilde{B}}, \tilde{\tilde{1}}\right)\right|
\end{equation}
\end{definition}
%
%
%

\section{Extended MABAC for group decision making based on IT2TrFNs}

The basic steps of MABAC method are: (i) Determine the weights of criteria (ii) Construction of initial decision matrix (iii) Computation of normalized decision matrix and weighted normalized decision matrix (iv) Finding Boarder approximation area(BAA) of each criteria functions for comparison of ideal and anti-ideal solutions  (v) Calculate the relative distance of each alternative’s from ideal solution using BAA (vi) Finally, rank the alternatives based on the overall closeness coefficient of each alternative’s to the ideal solution. Here, we extend MABAC method based on  interval type-2 trapezoidal fuzzy numbers(IT2TrFNs).\\
{\it Step 1. Determine the weighting of evaluation criteria}\\  
Construct the weighting matrix $W_{k}$ of the attributes of the $k$th decision-maker and construct the average weighting matrix $W$, respectively, shown as follows:
\begin{equation}
W_{k}=\left[\begin{array}{cccc}
\tilde{\tilde{w}}_{1}^k & \tilde{\tilde{w}}_{2}^k &...& \tilde{\tilde{w}}_{q}^k\\
\end{array}
\right]
\end{equation}
\begin{equation}
\bar{W}=(\tilde{\tilde{w}}_{j})_{1\times q}
\end{equation}
where $\tilde{\tilde{w}}_{j}=\left(\frac{\tilde{\tilde{w}}_{j}^1\oplus\tilde{\tilde{w}}_{j}^2\oplus\cdots\oplus\tilde{\tilde{w}}_{j}^k}{k}\right)$, $\tilde{\tilde{w}}_{j}$ is a IT2TrFN, $1\leq j\leq q$ and k denotes the number of decision makers.\\
{\it Step 2. Construct the IT2TrFNs performance/decision matrix }\\
Construct the decision matrix $D^k$ of the $k$-th decision-maker and construct the average decision matrix $D$, respectively, shown as follows
\begin{equation}
D^{k}=(\tilde{\tilde{a}}_{ij}^{k})_{p\times q}
\end{equation}
\begin{equation}
D=(\tilde{\tilde{a}}_{ij})_{p\times q}
\end{equation}
where $\tilde{\tilde{a}}_{ij}=\left(\frac{\tilde{\tilde{a}}_{ij}^1\oplus\tilde{\tilde{a}}_{ij}^2\oplus\cdots\oplus\tilde{\tilde{a}}_{ij}^K}{K}\right)$, $\tilde{\tilde{a}}_{ij}$ is a IT2TrFN, $1\leq i\leq p; 1\leq j\leq q$ and $k=1, 2, ..., K$ denotes the number of decision makers.\\ 
{\it Step 3. Normalization of the elements of the average decision matrix}\\
All relevant candidates' performance/property values, represented in the form of IT2TrFN, are first normalized and then weighted. For the normalization procedure, distinguishing must be made as to whether a material property is the larger the better (benefit type) or the lower the better (cost type). Let $\tilde{\tilde{a}}_{ij}$ and $\tilde{\tilde{n}}_{ij}$ denote the IT2TrFNs of $j$-th criteria of $i$-th alternative before and after normalization. The normalization procedure works as follows. For each property, compute the range. That is, $range_{j}=a_{j}^{+}-a_{j}^{-}$ where $a_{j+}^{U}=\max\limits_{1\leq i\leq p}\left(a_{ij4}^U\right)$ and $a_{j-}^{U}=\min\limits_{1\leq i\leq p}\left(a_{ij1}^U\right)$, $j=1, 2, ..., q$.\\
If attribute $j$ is of the benefit type then perform the following normalization operation
\begin{equation}
\begin{array}{cc}
\tilde{\tilde{n}}_{ij}=\left[\left(\frac{a_{ij1}^{U}-a_{j-}^{U}}{a_{j+}^{U}-a_{j-}^{U}}, \frac{a_{ij2}^{U}-a_{j-}^{U}}{a_{j+}^{U}-a_{j-}^{U}}, \frac{a_{ij3}^{U}-a_{j-}^{U}}{a_{j+}^{U}-a_{j-}^{U}}, \frac{a_{ij4}^{U}-a_{j-}^{U}}{a_{j+}^{U}-a_{j-}^{U}}; h_{A_{ij}}^{U}\right), \left(\frac{a_{ij1}^{L}-a_{j-}^{U}}{a_{j+}^{U}-a_{j-}^{U}}, \frac{a_{ij2}^{L}-a_{j-}^{U}}{a_{j+}^{U}-a_{j-}^{U}}, \frac{a_{ij3}^{L}-a_{j-}^{U}}{a_{j+}^{U}-a_{j-}^{U}}, \frac{a_{ij4}^{L}-a_{j-}^{U}}{a_{j+}^{U}-a_{j-}^{U}}; h_{A_{ij}}^{L}\right)\right]
\end{array}  
\end{equation}
If attribute $j$ is of the cost type then perform the following normalization operation
\begin{equation}
\begin{array}{cc}
\tilde{\tilde{n}}_{ij}=\left[\left(\frac{a_{j+}^{U}-a_{ij4}^{U}}{a_{j+}^{U}-a_{j-}^{U}}, \frac{a_{j+}^{U}-a_{ij3}^{U}}{a_{j+}^{U}-a_{j-}^{U}}, \frac{a_{j+}^{U}-a_{ij2}^{U}}{a_{j+}^{U}-a_{j-}^{U}}, \frac{a_{j+}^{U}-a_{ij1}^{U}}{a_{j+}^{U}-a_{j-}^{U}}; h_{A_{ij}}^{U}\right), \left(\frac{a_{j+}^{U}-a_{ij4}^{L}}{a_{j+}^{U}-a_{j-}^{U}}, \frac{a_{j+}^{U}-a_{ij3}^{L}}{a_{j+}^{U}-a_{j-}^{U}}, \frac{a_{j+}^{U}-a_{ij2}^{L}}{a_{j+}^{U}-a_{j-}^{U}}, \frac{a_{j+}^{U}-a_{ij1}^{L}}{a_{j+}^{U}-a_{j-}^{U}}; h_{A_{ij}}^{L}\right)\right]
\end{array}  
\end{equation}
{\it Step 4. Calculation of the weighted matrix}\\
The weighted matrix ($\tilde{\tilde{V}}$) is constructed as follows
\begin{equation}
\tilde{\tilde{V}}=(\tilde{\tilde{v}}_{ij})_{p\times q}
\end{equation}
where the elements of the weighted matrix $\tilde{\tilde{v}}_{ij}$ are calculated as follows
\begin{equation}
\tilde{\tilde{v}}_{ij}=\tilde{\tilde{w}}_{ij}\otimes(\tilde{\tilde{n}}_{ij}+\tilde{\tilde{1}})
\end{equation}
{\it Step 5. Calculation of Border Approximation Area (BAA)}\\
The border approximation area for each evaluation criterion is calculated using the geometric mean operator for IT2TrFNs [see Eqs.(9)-(11)] and defined by  
\begin{equation}
\tilde{\tilde{g}}_{j}=TIT2FGBM^{r, s} \Big(\tilde{\tilde{v}}_{1j}, \tilde{\tilde{v}}_{2j}, \cdots, \tilde{\tilde{v}}_{pj}\Big) 
\end{equation} 
After calculating BAA for each criteria function we now form the IT2TrFNs based border approximation area comparison matrix ($\tilde{\tilde{G}}$) as defined as
\begin{equation}
\tilde{\tilde{G}}=\left[\begin{array}{cccc}
\tilde{\tilde{g}}_{1} & \tilde{\tilde{g}}_{2} &...& \tilde{\tilde{g}}_{q}\\
\tilde{\tilde{g}}_{1} & \tilde{\tilde{g}}_{2} &...& \tilde{\tilde{g}}_{q}\\
. & . &...& .\\
\tilde{\tilde{g}}_{1} & \tilde{\tilde{g}}_{2} &...& \tilde{\tilde{g}}_{q}\\
\end{array}
\right]
\end{equation}
{\it Step 6. Calculation of distance matrix}\\
The rank-based distance operator [see Eq.(12)-(13)] for IT2TrFNs is used here to measure the ranked-based distances each performance in the weighted matrix $\tilde{\tilde{V}}$  are computed to form distance matrix ($Q$). Similarly, the ranked-based distance matrix $G$ of $\tilde{\tilde{V}}$ is calculated. These matrices are described as follows  
\begin{equation}
Q=(\tilde{\tilde{d}}_{ij})_{p\times q}
\end{equation}
where the elements of the distance matrix $d_{ij}$ are calculated as follows
\begin{equation}
d_{ij}=\left|R_{d}\left(\tilde{\tilde{v}}_{ij}, \tilde{\tilde{1}}\right)\right|~~(a~crisp~value)  
\end{equation}
\begin{equation}
G=(g_{ij})_{p\times q}
\end{equation}
where the elements of the distance matrix $d_{ij}$ are calculated as follows
\begin{equation}
g_{j}=\left|R_{d}\left(\tilde{\tilde{g}}_{j}, \tilde{\tilde{1}}\right)\right|~~(a~crisp~value)
\end{equation}
{\it Step 7. Determination of the ranking orders of all alternatives}\\
Now, boarder approximation area (BAA) value for each criteria function serves as reference point/benchmark value for criteria-wise performance of an alternative $A_{i}$. Each individual candidate will belong three different areas namely, the border approximation area ($G$), upper approximation area ($G^+$), and lower approximation area ($G^-$). The ideal alternative ($A_{i}^+$) can be found in the upper approximation area ($G^+$) whereas the lower approximation area ($G^-$) contains the anti-ideal alternative ($A_{i}^-$).  
\begin{equation}
A_{i}\in \left\{
   \begin{array}{ll}
   G^{+}, & \hbox{if $d_{ij}-g_{j}>0$} \\
   G, & \hbox{if $d_{ij}-g_{j}=0$} \\
   G^{-}, & \hbox{if $d_{ij}-g_{j}<0$} \\
  \end{array}
  \right.
\end{equation} 
Therefore, in order to select $A_{i}$ to be best compromised alternative (suitable candidate for the job), it is necessary for the candidate to be closest to $A_{i}^+$ and farthest from $A_{i}^-$. have as many criteria as possible belonging to the upper approximate area ($G^+$). The closeness coefficient  $S(A_{i})$ to the border approximation area for each alternative can be computed using the following equation.
\begin{equation}
S(A_{i})=\sum_{j=1}^{q} (d_{ij}-g_{j})
\end{equation}
The alternatives are ranked according to the decreasing value of $S(A_{i})$.
\section{Illustrative example}
In this section, we adopt one example from Chen and Lee \cite{chen10a} to illustrate the fuzzy multiple attributes group decision-making process of our proposed method. Assume that there are three decision-makers $DM1$, $DM2$ and $DM3$ of a software company to hire a system analysis engineer and assume that there are three alternatives/candidates for this job, say, $A1$, $A2$ and $A3$. The selection would be done based on five attributes namely, `Emotional Steadiness', `Oral Communication Skill', `Personality', `Past Experience', `Self-Confidence'. Let $X$ be the set of alternatives, where $X=\lbrace A1, A2, A3 \rbrace$, and let $F$ be the set of attributes, where $F=$ {\it$\lbrace$Emotional Steadiness; Oral Communication Skill; Personality; Past Experience; Self-Confidence$\rbrace$}.\\

Assume that the three decision makers $DM1$, $DM2$ and $DM3$ use the linguistic terms shown in Table 1 to represent the weights of the five attributes, respectively, as shown in Table 3. In Table 3, five benefit type (maximizing criteria) attributes are considered, including `Emotional Steadiness' (denoted by $C_{1}$), `Oral Communication Skill' (denoted by $C_{2}$), `Personality' (denoted by $C_{3}$), `Past Experience' (denoted by $C_{4}$) and `Self-Confidence' (denoted by $C_{5}$). Assume that the three decision-makers $DM1$, $DM2$ and $DM3$ use the linguistic terms shown in Table 2 to represent the evaluating values of the candidates with respect to different attributes, respectively, as shown in Table 5. The linguistic terms shown in Table 3 and 5 can be represented by the IT2TrFN ratings as shown in Tables 1 and 2 respectively.\\
Step 1: Based on Table 3 and Eq. (14-15), we can get the aggregated IT2TrFN weights for the decision attributes which are shown in Table 4.
\begin{center}
$\langle$ Table 3 and 4 about here $\rangle$
\end{center}
Step 2: Based on Table 5 and Eq. (16-17), we can construct the average initial decision matrix (Table 6).
\begin{center}
$\langle$ Table 5 and 6 about here $\rangle$
\end{center}
Step 3: For normalization it should be noted that in Table 2, the universe of discourse of the linguistic terms for the ratings is [0, 10]. Therefore, we normalize each entries in Table 6 according to Eq.(18-19), since all the attributes are maximizing type in this example. Data normalization is essential for all kinds of decision-making problems, to ensure we obtain  dimensionless  units,  from  heterogeneous  data  measurements,  which  can  be aggregated for rating and ranking decision alternatives.\\
Step 4: Based on Eq.(20-21), we can get the IT2TrFN-based weighted decision matrix $\tilde{\tilde{V}}$, shown as follows:
\begin{center}
$\langle$ Table 7 about here $\rangle$
\end{center}
Step 5. Using the geometric aggregation operation for interval valued numbers, the IT2TrFN-based border approximation area (BAA) matrix (Table 8) for each evaluation criterion is computed according to Eq.(22).
\begin{center}
$\langle$ Table 8 about here $\rangle$
\end{center}
Step 6. Rank-based distance matrices (Table 9 and 10) of weighted matrix and BAA are calculated using Eq.(12-13) in order to compare the attribute-wise performances of each candidate.
\begin{center}
$\langle$ Table 9 and 10 about here $\rangle$
\end{center}
Step 7. Finally, we can verify each candidate's strength and weakness in every attribute by subtracting $G$ from $Q$ as defined above in Eq.(24) and (26). The final score values according to Eq.(29) of the candidates and final rank of the candidates is given in Table 11.    
\begin{center}
$\langle$ Table 11 about here $\rangle$
\end{center}
\section{Comparison and Discussion}
In order to verify the validity of our proposed method, we perform a comparison of our proposed method with two other previous methods including TOPSIS by Chen and Lee \cite{chen10a} and Gong et al. \cite{gong15} which also deal with IT2TrFNs. The results are shown as follows:
\begin{center}
$\langle$ Table 12 about here $\rangle$
\end{center}
Now, let us observe few things from Table 11. Candidate $A1$ has better `Past Experience' than $A2$ and $A3$ where as $A2$ performed best in all others attributes. So, it is clear that no single candidate is best in all the attributes. So, we proceed for a compromised solution. It is also clear from Table 12 that the three methods have the similar results. Note that $A2$ is best compromised alternative/candidate according to all three models under fixed preference of attribute weights. This shows the method we proposed in this paper is reasonable.\\

The main reason of using MABAC method is the simple computation procedure and the stability (consistency) of solution (Pamu{\v{c}}ar and {\'C}irovi{\'c} \cite{pamucar15}; Xue et al. \cite{xue16}). The MABAC method is a particularly pragmatic and reliable tool for rational decision making. One more benefit of this method is that it enables us to visualize of performance and assessment of individual candidates as per each criteria and vice versa (Table 11). From the distance matrix Table 11,  one can directly conclude whether an alternative performs better than the ideal person or not. In other words, this particular method enables us to understand easily the strength and weakness of a candidate compared to others. But TOPSIS [20] and VIKOR [18] methods do not produce such a direct observation.

\section{Conclusions}
This study proposes an extended MABAC for multi-attribute decision making model based on IT2TrFNs to facilitate a more precise analysis of the alternatives, considering several criteria in fuzzy environment. Different relative weights of attributes is more realistic in many practical MADM problems, especially in complex and uncertain environments. MABAC possess simple computation procedure and the stability (consistency) of solution. Here, we utilize IT2TrFN-based MABAC to evaluate and select the most suitable candidate for a software company which is heading to hire a system analysis engineer based on few attributes. Finally, the proposed method is validated comparing with other two methods from literature. The proposed method could be applied to other MADM problems such as supplier selection, project portfolio selection etc. In future applications of the IT2TrFN-based MABAC method it would be interesting to combine with various methods for determining the weights of attributes (e.g., Entropy, AHP, DEMATEL, ANP etc).

\subsubsection*{Acknowledgment.} The first author would like to thank Department of Science and Technology, New Delhi, India, for their help and supports in this research work under INSPIRE program with sanction order number: DST/INSPIRE Fellowship/2013/544.

\begin{table}[] 
\centering
\caption{Linguistic terms and related TrIT2FN ratings for criteria priorities}
\begin{tabular}{l l l l}
\hline
Linguistic terms&& TrIT2FN Ratings\\
\hline									
Very Low(VL)& &      [(0,  0,	 0,  .1;    1), (0,      0,  0, .05; 0.9)]\\
Low (L)& & 	         [(0,  .1,	.1,  .3;	1),	(.05,   .1,	.1,  2; 0.9)]\\
Medium Low (ML)& &   [(.1, .3,	.3,  .5;	1),	(.2,	.3,	.3,	 4;	0.9)]\\
Medium (M)& &	     [(.3, .5,	.5,  .7;	1),	(.4,	.5,	.5,	 6;	0.9)]\\
Medium High (MH)& &  [(.5, .7,	.7,  .9;	1),	(.6,	.7,	.7,	 8;	0.9)]\\
High (H)& &	         [(.7, .9,	.9,   1;	1),	(.8,	.9,	.9,	.95; 0.9)]\\
Very High (VH)& &	 [(.9,	1,   1,   1;	1),	(.95,    1,	 1,	 1;	0.9)]\\
\hline		
\end{tabular}
\end{table}
\begin{table}[] 
\centering
\caption{Linguistic terms and related TrIT2FN ratings for evaluating alternatives} 
\begin{tabular}{l l l l}
\hline
Linguistic terms&& TrIT2FN Ratings\\
\hline									
Very Poor(VP)& &    [(0, 0,	0, 1; 1), (0, 0, 0, 0.5; 0.9)]\\
Poor (P)& & 	      [(0,	1,	1,	3;	1),	(0.5,   1,	1, 2; 0.9)]\\
Medium Poor (MP)& &  [(1,	3,	3,	5;	1),	(2,	3,	3,	4;	0.9)]\\
Fair (F)& &	      [(3,	5,	5,	7;	1),	(4,	5,	5,	6;	0.9)]\\
Medium Good (MG)& &  [(5,	7,	7,	9;	1),	(6,	7,	7,	8;	0.9)]\\
Good (G)& &	      [(7,	9,	9,	10;	1),	(8,	9,	9,	9.5; 0.9)]\\
Very Good (VG)& &	  [(9,	10,	10,	10;	1),	(9.5,10	10,	10;	0.9)]\\
\hline		
\end{tabular}
\end{table}
\begin{table}[] 
\centering
\caption{TrIT2FN-based weights for attribute/criteria set}
\begin{tabular}{c c c c}
\hline
&	DM1&	DM2&	DM3\\
\hline
$C_{1}$&	H&	VH&	MH\\
$C_{2}$&	VH&	VH&	VH\\
$C_{3}$&	VH&	H&	H\\
$C_{4}$&	VH&	VH&	VH\\
$C_{5}$&	M&	MH&	MH\\
\hline		
\end{tabular}
\end{table}
\begin{table}[] 
\centering
\caption{Aggregated weights for attribute/criteria set} 
\begin{tabular}{c c c c c c c c c c c}
\hline
$C_{1}$&	[(0.70,&	0.87,&	0.87,&	0.97;&	1),&	(0.70,&	0.87,&	0.87,&	0.97;&	0.9)]\\
$C_{2}$&	[(0.90,&	1.00,&	1.00,&	1.00;&	1),&	(0.90,&	1.00,&	1.00,&	1.00;&	0.9)]\\
$C_{3}$&	[(0.77,&	0.93,&	0.93,&	1.00;&	1),&	(0.77,&	0.93,&	0.93,&	1.00;&	0.9)]\\
$C_{4}$&	[(0.90,&	1.00,&	1.00,&	1.00;&	1),&	(0.90,&	1.00,&	1.00,&	1.00;&	0.9)]\\
$C_{5}$&	[(0.43,&	0.63,&	0.63,&	0.83;&	1),&	(0.43,&	0.63,&	0.63,&	0.83;&	0.9)]\\
\hline		
\end{tabular}
\end{table}
\begin{table}[] 
\centering
\caption{Experts' ratings based on TrIT2FNs for alternatives} 
\begin{tabular}{c c c c c c c c c c c c c c c c}
\hline
& $C_{1}$& & & $C_{2}$& & & $C_{3}$& & & $C_{4}$& & & $C_{5}$& & \\									
\hline		
& DM1& DM2& DM3& DM1& DM2& DM3& DM1& DM2& DM3& DM1& DM2& DM3& DM1& DM2& DM3\\
\hline		
A1&	MG&	G&	MG&	G&	MG&	F&	F&	G&	G&	VG&	G&	VG&	F&	F&	F\\
A2&	G&	G&	MG&	VG&	VG&	VG&	VG&	VG&	G&	VG&	VG&	VG&	VG&	MG&	G\\
A3&	VG&	G&	F&	G&	G&	VG&	G&	MG&	VG&	G&	VG&	MG&	MG&	G&	MG\\
\hline		
\end{tabular}
\end{table}
\begin{table}[] 
\centering
\caption{TrIT2FN-based aggregated initial decision matrix}
\begin{tabular}{c c c c c c c c c c c}
\hline
& $C_{1}$& & & & & & & & &\\									
\hline					
A1&	[(5.67,&	7.67,&	7.67,&	9.33;&	1),&	(6.67,&	7.67,&	7.67,&	8.50;&	0.9)]\\
A2&	[(6.33,&	8.33,&	8.33,&	9.67;&	1),&	(7.33,&	8.33,&	8.33,&	9.00;&	0.9)]\\
A3&	[(6.33,&	8.00,&	8.00,&	9.00;&	1),&	(7.17,&	8.00,&	8.00,&	8.50;&	0.9)]\\
\hline
& $C_{2}$& & & & & & & & &\\									
\hline									
A1&	[(5.00,&	7.00,&	7.00,&	8.67;&	1),&	(6.00,&	7.00,&	7.00,&	7.83;&	0.9)]\\
A2&	[(9.00,&	10.00,&	10.00,&	10.00;&	1),&	(9.50,&	10.00,&	10.00,&	10.00;&	0.9)]\\
A3&	[(7.00,&	8.67,&	8.67,&	9.67;&	1),&	(7.83,&	8.67,&	8.67,&	9.17;&	0.9)]\\
\hline
& $C_{3}$& & & & & & & & &\\									
\hline									
A1&	[(5.67,&	7.67,&	7.67,&	9.00;&	1),&	(6.67,&	7.67,&	7.67,&	8.33;&	0.9)]\\
A2&	[(9.00,&	10.00,&	10.00,&	10.00;&	1),&	(9.50,&	10.00,&	10.00,&	10.00;&	0.9)]\\
A3&	[(7.00,&	8.67,&	8.67,&	9.67;&	1),&	(7.83,&	8.67,&	8.67,&	9.17;&	0.9)]\\
\hline
& $C_{4}$& & & & & & & & &\\									
\hline									
A1&	[(8.33,&	9.67,&	9.67,&	10.00;&	1),&	(9.00,&	9.67,&	9.67,&	9.83;&	0.9)]\\
A2&	[(7.67,&	9.00,&	9.00,&	9.67;&	1),&	(8.33,&	9.00,&	9.00,&	9.33;&	0.9)]\\
A3&	[(7.00,&	8.67,&	8.67,&	9.67;&	1),&	(7.83,&	8.67,&	8.67,&	9.17;&	0.9)]\\
\hline
& $C_{5}$& & & & & & & & &\\									
\hline								
A1&	[(3.00,&	5.00,&	5.00,&	7.00;&	1),&	(4.00,&	5.00,&	5.00,&	6.00;&	0.9)]\\
A2&	[(6.33,&	8.00,&	8.00,&	9.33;&	1),&	(7.17,&	8.00,&	8.00,&	8.67;&	0.9)]\\
A3&	[(6.33,&	8.33,&	8.33,&	9.67;&	1),&	(7.33,&	8.33,&	8.33,&	9.00;&	0.9)]\\
\hline		
\end{tabular}
\end{table}
\begin{table}[] 
\centering
\caption{TrIT2FN-based weighted decision matrix}  
\begin{tabular}{c c c c c c c c c c c}
\hline
& $C_{1}$& & & & & & & & &\\									
\hline					
A1&	[(0.70,&	1.30,&	1.30,&	2.00;&	1),&	(0.88,&	1.30,&	1.30,&	1.94;&	0.9)]\\
A2&	[(0.82,&	1.44,&	1.44,&	1.89;&	1),&	(0.99,&	1.44,&	1.44,&	1.78;&	0.9)]\\
A3&	[(0.82,&	1.37,&	1.37,&	1.89;&	1),&	(0.96,&	1.37,&	1.37,&	1.72;&	0.9)]\\
\hline
& $C_{2}$& & & & & & & & &\\									
\hline									
A1&	[(0.90,&	1.40,&	1.40,&	1.73;&	1),&	(1.08,&	1.40,&	1.40,&	1.57;&	0.9)]\\
A2&	[(1.62,&	2.00,&	2.00,&	2.00;&	1),&	(1.71,&	2.00,&	2.00,&	2.00;&	0.9)]\\
A3&	[(1.26,&	1.73,&	1.73,&	1.93;&	1),&	(1.41,&	1.73,&	1.73,&	1.83;&	0.9)]\\
\hline
& $C_{3}$& & & & & & & & &\\									
\hline									
A1&	[(0.77,&	1.36,&	1.36,&	1.77;&	1),&	(0.95,&	1.36,&	1.36,&	1.62;&	0.9)]\\
A2&	[(1.36,&	1.87,&	1.87,&	2.00;&	1),&	(1.45,&	1.86,&	1.86,&	2.00;&	0.9)]\\
A3&	[(1.00,&	1.58,&	1.58,&	1.92;&	1),&	(1.16,&	1.57,&	1.57,&	1.81;&	0.9)]\\
\hline
& $C_{4}$& & & & & & & & &\\									
\hline									
A1&	[(1.30,&	1.89,&	1.89,&	2.00;&	1),&	(1.50,&	1.89,&	1.89,&	1.94;&	0.9)]\\
A2&	[(1.10,&	1.67,&	1.67,&	1.89;&	1),&	(1.30,&	1.67,&	1.67,&	1.78;&	0.9)]\\
A3&	[(0.90,&	1.56,&	1.56,&	1.89;&	1),&	(1.15,&	1.56,&	1.56,&	1.72;&	0.9)]\\
\hline
& $C_{5}$& & & & & & & & &\\									
\hline									
A1&	[(0.43,&	0.82,&	0.82,&	1.33;&	1),&	(0.49,&	0.82,&	0.82,&	1.20;&	0.9)]\\
A2&	[(0.65,&	1.11,&	1.11,&	1.63;&	1),&	(0.70,&	1.10,&	1.10,&	1.54;&	0.9)]\\
A3&	[(0.65,&	1.14,&	1.14,&	1.67;&	1),&	(0.71,&	1.13,&	1.13,&	1.58;&	0.9)]\\
\hline		
\end{tabular}
\end{table}
\begin{table}[] 
\centering
\caption{TrIT2FN-based BAA comparison matrix} 
\begin{tabular}{c c c c c c c c c c c}
\hline
$C_{1}$&	[(0.78,&	1.37,&	1.37,&	1.85;&	1),&	(0.94,&	1.37,&	1.37,&	1.70;&	0.90)]\\
$C_{2}$&	[(1.22,&	1.69,&	1.69,&	1.89;&	1),&	(1.38,&	1.69,&	1.69,&	1.79;&	0.90)]\\
$C_{3}$&	[(1.01,&	1.59,&	1.59,&	1.89;&	1),&	(1.17,&	1.58,&	1.58,&	1.80;&	0.90)]\\
$C_{4}$&	[(1.09,&	1.70,&	1.70,&	1.93;&	1),&	(1.31,&	1.70,&	1.70,&	1.81;&	0.90)]\\
$C_{5}$&	[(0.57,&	1.01,&	1.01,&	1.53;&	1),&	(0.63,&	1.01,&	1.01,&	1.43;&	0.90)]\\
\hline		
\end{tabular}
\end{table}
\begin{table}[] 
\centering
\caption{Rank-based distance matrix of weighted matrix} 
\begin{tabular}{c c c c c c}
\hline
&	$C_{1}$& $C_{2}$& $C_{3}$& $C_{4}$& $C_{5}$\\
\hline	
A1&	1.75&	1.94&	1.85&	2.90&	0.75\\
A2&	2.03&	3.11&	2.81&	2.46&	1.32\\
A3&	1.87&	2.60&	2.27&	2.26&	1.39\\
\hline		
\end{tabular}
\end{table}
\begin{table}[] 
\centering
\caption{Rank-based distance matrix of BAA}
\begin{tabular}{c c c c c}
\hline
$C_{1}$& $C_{2}$& $C_{3}$& $C_{4}$& $C_{5}$\\
\hline	
1.88&	2.52&	2.29&	2.53&	1.13\\
1.88&	2.52&	2.29&	2.53&	1.13\\
1.88&	2.52&	2.29&	2.53&	1.13\\
\hline		
\end{tabular}
\end{table}
\begin{table}[] 
\centering
\caption{Rank-based distance matrix of BAA}
\begin{tabular}{c |c c c c c |c| c}
\hline
& $C_{1}$& $C_{2}$& $C_{3}$& $C_{4}$& $C_{5}$& Final Score $S(A_{i})$ & Ranking\\
\hline	
A1&	-0.13&	-0.58&	-0.44&	0.37&	-0.38&	-1.16&	3\\
A2&	0.15&	0.59&	0.52&	-0.07&	0.19&	1.39&	1\\
A3&	-0.01&	0.08&	-0.02&	-0.27&	0.25&	0.03&	2\\
\hline		
\end{tabular}
\end{table}

\begin{table}[] 
\centering
\caption{Comparison with other methods} 
\begin{tabular}{c c c c}
\hline
& Ranking order by & Ranking order by & Ranking order by \\
& Chen and Lee (2010)& Gong (2014)& Proposed Model\\
\hline	
A1&	3& 3& 3\\
A2&	1& 1& 1\\
A3&	2& 2& 2\\
\hline		
\end{tabular}
\end{table}

\end{document}